\documentclass[runningheads]{llncs}
\usepackage[T1]{fontenc}
\usepackage{graphicx}
\usepackage{tikz}
\usepackage{comment}
\usepackage{amsmath,amssymb} 
\usepackage{color}

\usepackage[accsupp]{axessibility}  

\usepackage{booktabs}
\usepackage{multirow}
\usepackage{multicol}
\usepackage{arydshln}
\usepackage[misc]{ifsym}

\begin{document}
\title{ITTR: Unpaired Image-to-Image Translation with Transformers}
%

\author{
Wanfeng Zheng\inst{1,2}\and
Qiang Li\inst{2}\thanks{Corresponding author} \and
Guoxin Zhang\inst{2}\and
Pengfei Wan\inst{2}\and
Zhongyuan Wang\inst{2}
}


%
\authorrunning{W.F. Zheng, Q. Li et al.}
%
\institute{
Beijing University of Posts and Telecommunications \\
\email{zhengwanfeng@bupt.edu.cn} \and
Kuaishou Technology\\
\email{\{liqiang03,zhangguoxin,wanpengfei,wangzhongyuan\}@kuaishou.com}}

\maketitle              

\begin{figure}[h!]
  \centering
  \includegraphics[width=\linewidth]{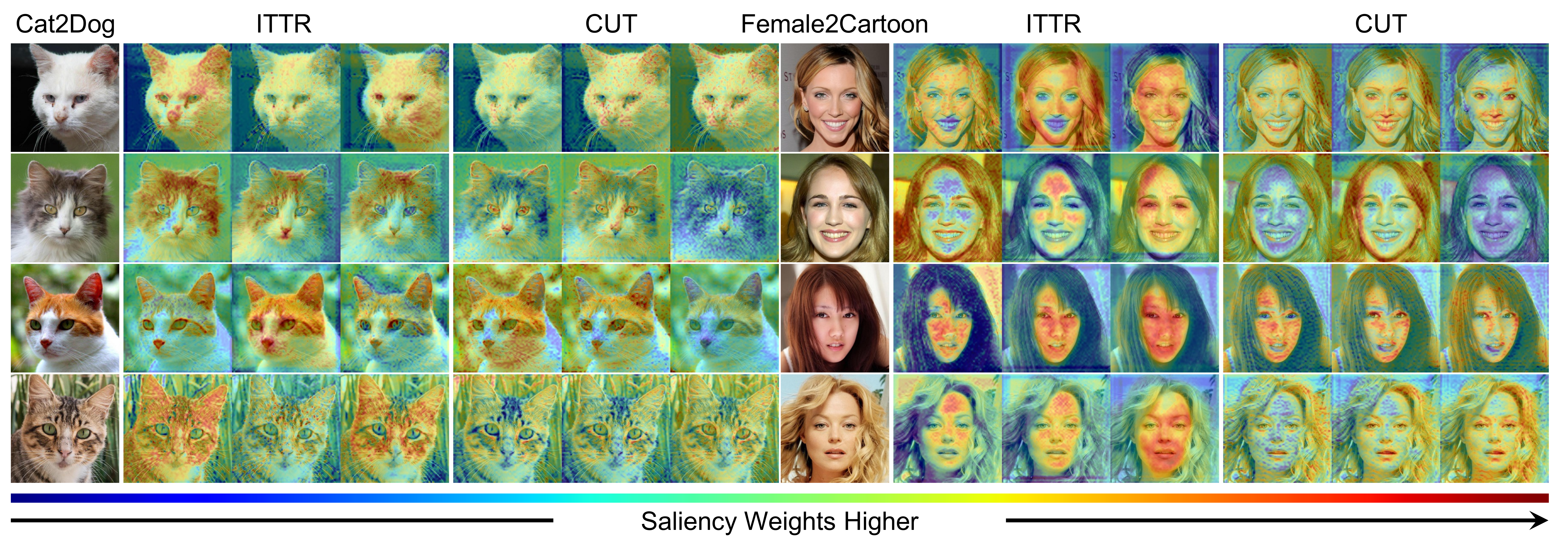}
  \caption{
    Comparison between ITTR and CUT \cite{cut} with the Grad-CAM \cite{gcam} visualization extracted from the first three blocks in the body of the generator.
    The bottom colorbar displays the relative saliency weight value of Grad-CAM, of which a red area indicates more contribution to the image-to-image translation and a blue one the opposite.
    By comparison, the transformer-based ITTR is more likely to capture instance-level contextual information than the CNN-based CUT.
  }
  \label{fig:gcam}
\end{figure}

\begin{abstract}
Unpaired image-to-image translation is to translate an image from a source domain to a target domain without paired training data.
By utilizing CNN in extracting local semantics, various techniques have been developed to improve the translation performance.
However, CNN-based generators lack the ability to capture long-range dependency to well exploit global semantics.
Recently, Vision Transformers have been widely investigated for recognition tasks.
Though appealing, it is inappropriate to simply transfer a recognition-based vision transformer to image-to-image translation due to the generation difficulty and the computation limitation.
In this paper, we propose an effective and efficient architecture for unpaired Image-to-Image Translation with Transformers (ITTR). 
It has two main designs: 1) hybrid perception block (HPB) for token mixing from different receptive fields to utilize global semantics; 2) dual pruned self-attention (DPSA) to sharply reduce the computational complexity.
Our ITTR outperforms the state-of-the-arts for unpaired image-to-image translation on six benchmark datasets.

\end{abstract}
%

\begin{figure}[t]
  \centering
  \includegraphics[width=\linewidth]{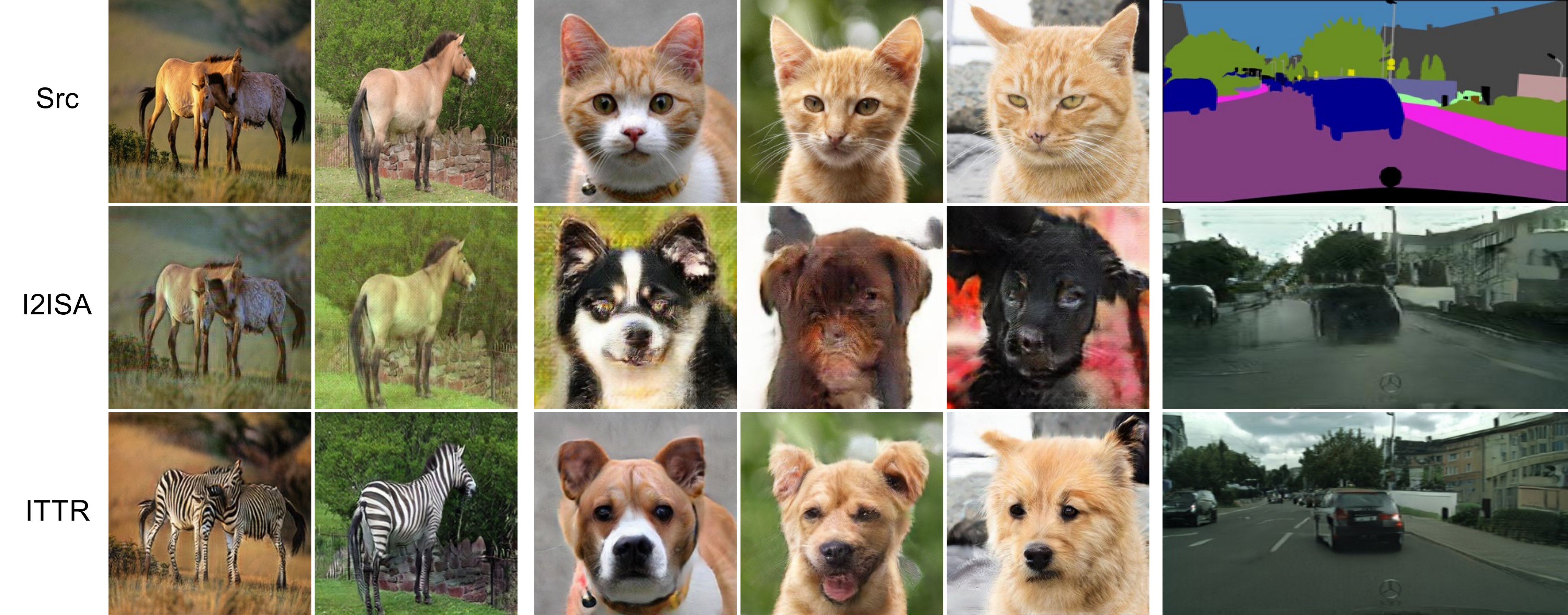}

  \caption{
   Qualitative comparison of I2ISA \cite{i2i_sa} and ITTR. 
   Though both of the methods have integrated self-attention into their network architecture, translation performance of the former is far behind the latter.
  }
  \label{fig:i2isa}
\end{figure}

\section{Introduction}


Unpaired image-to-image translation is to translate an image from the source domain to the target domain without paired training data.
The translation result should be constructed of content from the source domain and style from the target domain.
Most of the important unpaired image-to-image translation approaches \cite{cyclegan,cut,lsesim} pay attention to the development of training strategies. 
However, the generator architecture is still based on convolutional neural networks (CNN).
Though CNN is effective in extracting local semantics, it lacks the ability to capture long-range dependency which can be observed in Fig. \ref{fig:gcam}.

To enrich CNN's capability, a previous work I2ISA \cite{i2i_sa} has attempted to adopt non-local block \cite{non-local} into consideration.
In particular, motivated by SAGAN \cite{sagan}, I2ISA \cite{i2i_sa} embedded a non-local block into their generator architecture to enhance the performance of unpaired image-to-image translation. 
Though effective at large, the translation performance of I2ISA is still limited to some extent as shown in Fig. \ref{fig:i2isa}. 
One possible reason for this result is that I2ISA only has one single non-local block in the decoder of the generator which still lacks the ability to adequately exploit global semantics.

To fully capture long-range dependency, a potential solution is to introduce transformer-based generator architecture for unpaired image-to-image translation.
In fact, a vision transformer can stack with several multi-head self-attention (MHSA) blocks to continuously capture long-range dependency.
However, a recognition-based transformer is unsuitable for a generation task like unpaired image-to-image translation of which visual quality and semantic consistency are required.
Besides, a recognition-based transformer also suffers from huge computational cost and memory consumption due to the algorithmic complexity of MHSA when the input image resolution is high.
Therefore, novel designs are needed to deal with the generation difficulty and the computation limitation.

\begin{figure}[t!]
  \centering
  \includegraphics[width=\linewidth]{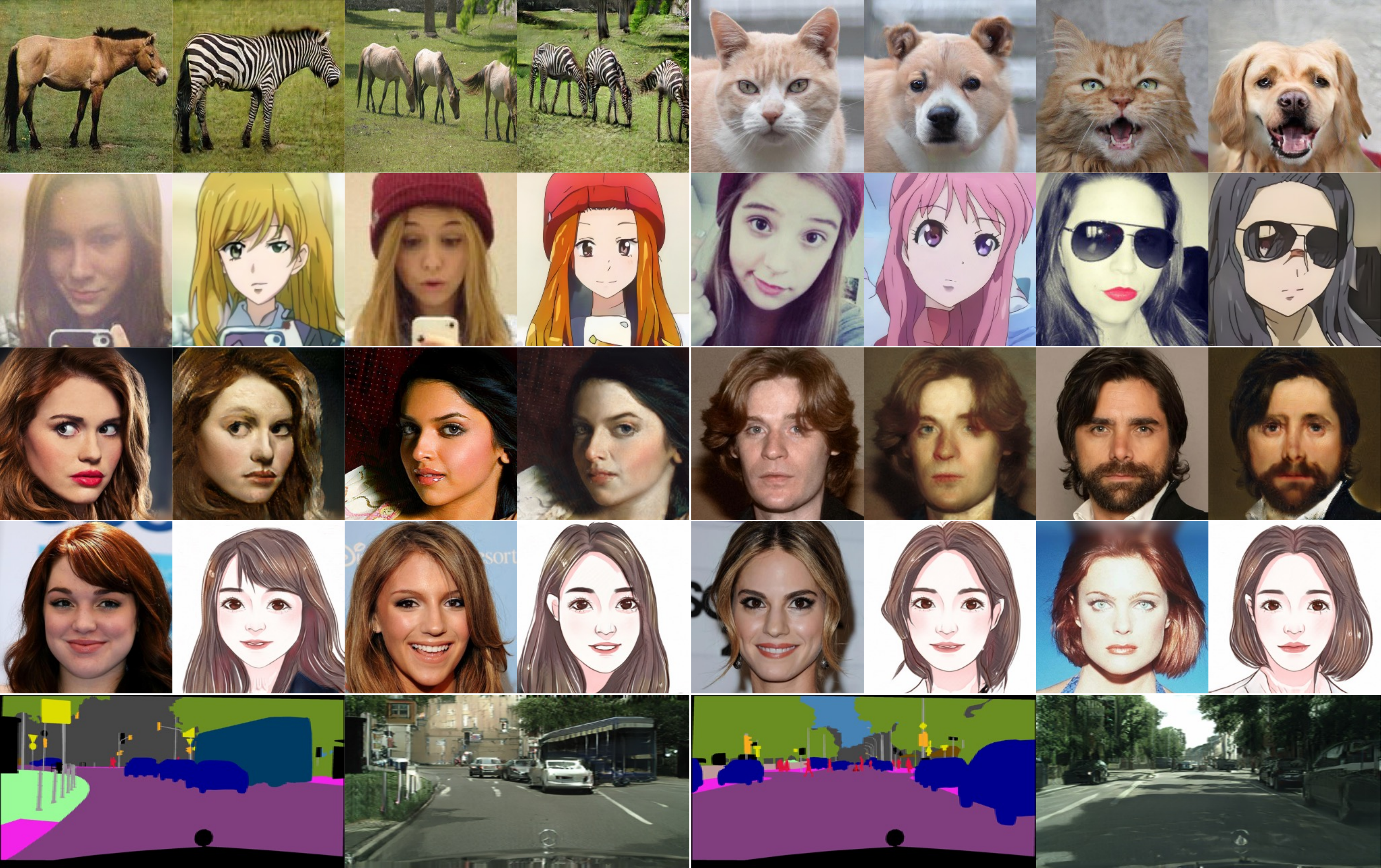}

  \caption{
  Examples of translation results produced by ITTR. From left to right, top to bottom: Horse $\rightarrow$ Zebra, Cat $\rightarrow$ Dog, Selfie $\rightarrow$ Anime, Face $\rightarrow$ MetFace, Female $\rightarrow$ Cartoon, Cityscapes. }
  \label{fig:demos}
\end{figure}


In this paper, we propose an effective and efficient architecture for unpaired Image-to-Image Translation with Transformers (ITTR). 
It has two targeted designs.
Firstly, we propose the hybrid perception block (HPB) for spatially token mixing from different receptive fields which aims to solve generation difficulties. 
HPB is able to extract short-range and long-range contextual information by a depth-wise convolution branch and a self-attention branch respectively.
Secondly, we propose dual pruned self-attention (DPSA) to reduce the complexity of MHSA.
Before calculating the attention map, DPSA evaluates the contribution of tokens along rows and columns.
Then, the less contributed tokens are pruned so that the complexity of the attention map is sharply reduced.

To demonstrate the effectiveness of our approach, we conduct both qualitative and quantitative experiments for unpaired image-to-image translation on six benchmark datasets including Horse $\rightarrow$ Zebra \cite{cyclegan}, Cityscapes \cite{cityscapes}, Cat $\rightarrow $ Dog \cite{stargan2}, Face $\rightarrow$ Metface \cite{metface}, Female $\rightarrow$ Cartoon \cite{cartoon} and Selfie $\rightarrow$ Anime \cite{ugatit}.
Fig. \ref{fig:demos} shows some of the qualitative results produced by ITTR. 
By comparison with the state-of-the-art methods, ITTR achieves better performance both qualitatively and quantitatively. 

Our contributions are summarized as follows:
\begin{itemize}
    \item A transformer-based architecture named ITTR is proposed for the task of unpaired image-to-image translation.
    
    \item Hybrid perception block (HPB) is designed to capture contextual information from different perceptions both locally and globally.
    
    \item Dual pruned self-attention (DPSA) is proposed to reduce the computational complexity and memory consumption while maintaining the performance.

    \item Experiments on six datasets validate the effectiveness of our approach both qualitatively and quantitatively. Particularly, our method outperforms the state-of-the-arts in terms of both FID \cite{fid} and DRN-Score \cite{cut}.
\end{itemize}

\section{Related Work}

In this section, we briefly review related works about image-to-image translation, vision transformers and generative vision transformers.

\subsection{Image-to-Image Translation} 

Image-to-image translation is an image generation task that converts image to the target domain with a new style while maintaining the content from the source domain. 
Development of paired image-to-image translation \cite{pix2pix,pix2pixhd} has been limited because of the difficulty of datasets preparation. 
Instead, unpaired image-to-image translation \cite{cyclegan,discogan} has achieved some success due to the proposal of reconstruction objective based on cycle consistency. 
This kind of bi-directional methods include CycleGAN \cite{cyclegan}, DiscoGAN \cite{discogan}, U-GAT-IT \cite{ugatit} etc. 
However, during the training process, these approaches can take up a lot of GPU memory because both sides are trained simultaneously. 
To handle this issue, one-sided unpaired image-to-image translation \cite{distancegan,gcgan} has been studied by leveraging the geometry consistency. 
Recently, some contrastive learning based one-sided translation approaches \cite{cut,negcut,lsesim} have shown dominance by exploiting contrastive consistency though at much lower GPU memory cost.
In our work, we employ the one-sided unpaired translation paradigm.

\subsection{Vision Transformers} 
Based on the original transformer \cite{transformer} designed for natural language processing tasks, Vision Transformer (ViT) \cite{vit} re-formulates computer vision tasks in terms of patch tokens.
Following ViT, a lot of transformer variants \cite{twins,hrformer,deit,crossformer} have been proposed to improve the effectiveness and efficiency. 
For example, Swin Transformer \cite{swin} has made great performance on several vision tasks with its shifted window mechanism. 
Some works like LeViT \cite{levit}, Mobile-Former \cite{mobileformer} and MobileViT \cite{mobilevit} focus on development of lightweight vision transformers.
Hierarchical vision transformers \cite{t2t,cvt,pit} enable efficient application of ViTs to dense prediction tasks. 
PVT \cite{pvtv1} has proposed spatial reduction attention with a convolution layer for down-sampling on Key and Value to reduce computation cost and memory consumption.
Later, the spatial reduction convolution layer has been replaced by a depth-wise convolution layer for further lightweight in CMT \cite{cmt}. PVTv2 \cite{pvtv2} has also adopted convolution layers for overlapping patch embedding and feed-forward computation.
Distinct from present works made for image recognition tasks, our work aims to develop a generative vision transformer for unpaired image-to-image translation.

\subsection{Generative Vision Transformers} 

Generative adversarial networks \cite{gan} based on CNN have been widely used in various tasks. 
Due to the success of ViT, several works have investigated unconditional generative transformers.
For example, TransGAN \cite{transgan} and ViTGAN \cite{vitgan} consider the task of image generation based on the pure transformer architecture.
Styleformer \cite{styleformer} has a transformer-based generator with style modulation and demodulation operation in MHSA. 
StyleSwin \cite{styleswin} has a StyleGAN \cite{stylegan,stylegan2} liked generator architecture for high-resolution image generation.
Different from StyleGAN, TokenGAN \cite{tokengan} uses tokens instead of latent codes to control the generated image style.
HiT \cite{hit} with its multi-axis self-attention is also proposed for high-resolution image generation.
Another line of research focuses on designing conditional generative transformers. For example,
Paint Transformer \cite{paintformer} makes feed-forward neural painting with stroke prediction. 
StyTr\^{}2 \cite{styletr2} designs two transformer encoders for arbitrary image stylization.
Different from general transformer, GANformer \cite{ganformer} has proposed simplex and duplex attention for image generation and replaced general FFN with ResNet \cite{resnet} block. GANformer2 \cite{ganformer2} has disassembled image synthesis into a planning stage and a execution stage for more explicitly scene generation. 
Our work belongs to conditional generative transformers and targets the general task of image-to-image translation, while most of the above methods are either unconditional or task-specific.

\begin{figure}[t]
  \centering
  \includegraphics[width=\linewidth]{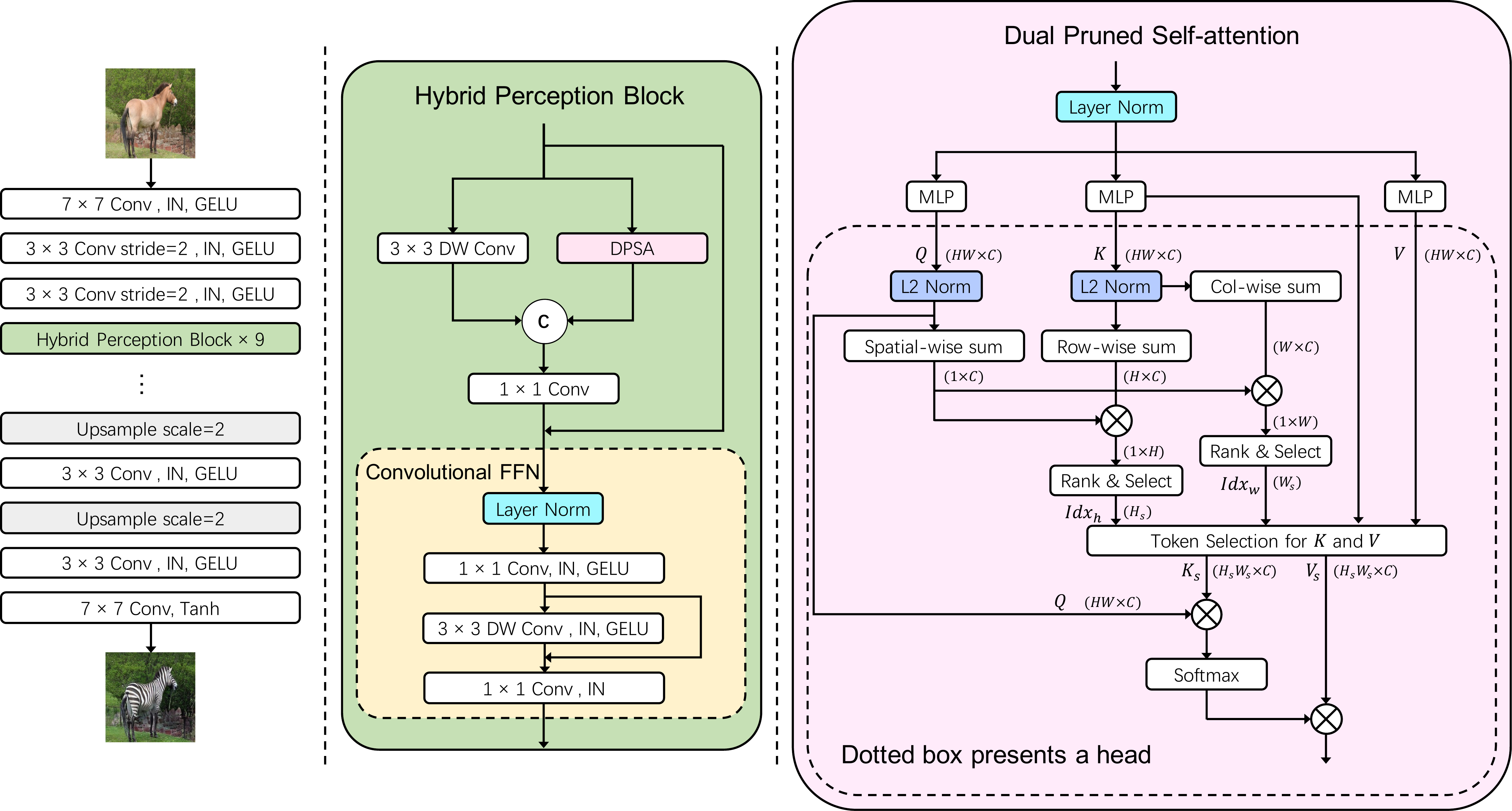}

  \caption{
    The architecture of ITTR.
    From left to right: the overall architecture of ITTR, the architecture of hybrid perception block (HPB), dual pruned self-attention (DPSA).
    ``Conv'' means convolution. ``IN'' is the abbreviation of instance normalization. ``DW'' means depth-wise. ``C'' in the circle represents concatenation.  
    ``L2 Norm'' is token-wise L2 normalization.
    The spatial-wise sum is to calculate the summation of all the tokens in $Q$. 
    Row-wise or Col-wise sum is to calculate the summation of tokens in the same row or column in $K$.
    Token selection is to select rows and columns from a matrix with indexes for referring. 
    ``$\times$'' in the circle represents matrix multiplication. 
    The dotted box in right represents operations made in a single head of DPSA.
  }
  \label{fig:arch}
\end{figure}

\section{Method}
In this section, we first introduce some preliminaries on the general architecture of a vision transformer. Then, we illustrate the detailed designs of ITTR in terms of the overall architecture and the building blocks, hybrid perception block and dual pruned self-attention, as shown in Fig. \ref{fig:arch}. Finally, we present the learning objectives for completeness.



\subsection{Preliminaries}

Transformer \cite{transformer} was first proposed as a network architecture for natural language processing (NLP) tasks.
Later, ViT \cite{vit} has introduced transformer-based architecture for vision task like image classification.
ViT is a pure transformer-based network consisted by MHSA and FFN. 
In ViT, patches of a 2D image $I \in \mathbb{R}^{H \times W \times 3}$ are embedded into tokens $X \in \mathbb{R}^{N \times C}$. Then the tokens are split into $N_{h}$ heads $X \in \mathbb{R}^{N_{h} \times N \times \frac{C}{N_{h}}}$. In each head, $X_i \in \mathbb{R}^{N \times D_{h}}$, $i \in \{1,2...N_{h}\}$, $D_{h}=\frac{C}{N_{h}}$, is used to calculated $Q_{i}$, $K_{i}$ and $V_{i}$ by linear layers.
\begin{equation}
  Q_{i},\, K_{i},\, V_{i} = X_{i}W_q,\, X_{i}W_k,\, X_{i}W_v, \ i \in \{1,2...N_{h}\}.
  \label{eq:qkv}
\end{equation}

Here, $W_q,\,W_k,\, W_v \in\mathbb{R}^{C \times C}$ represent learnable parameter matrixes corresponding to Query, Key and Value. Then, self-attention is calculated with $Q_{i}$, $K_{i}$ and $V_{i}$.
\begin{equation}
  \text{Attention}_i(X) = \text{Softmax}(\frac{Q_{i}K_{i}^{T}}{\sqrt{D_{h}}})V_{i}.
  \label{eq:att}
\end{equation}

Results from each head are concatenated and fed into a linear layer with bias. 

\begin{equation}
  \text{MHSA}(X) = \text{concat}_{i=1}^{N_{h}}[\text{Attention}_{i}(X_{i})]W+b.
  \label{eq:mhsa}
\end{equation}

Then the concatenated feature map is entered into the feed-forward network (FFN).
\begin{equation}
  \text{FFN}(X) = \text{MLP}(\text{MLP}(X)).
  \label{eq:ffn}
\end{equation}

Some recent works \cite{pvtv2,cmt} have inserted a depth-wise convolution layer into FFN. This variant is named inverted feed-forward network (IFFN) because it is similar to the inverted residual block in MobileNetV2 \cite{mobilev2}.
\begin{equation}
  \text{IFFN}(X) = \text{MLP}(\text{DWConv}(\text{MLP}(X))).
  \label{eq:invertedffn}
\end{equation}

\subsection{Overall Architecture}


As shown in Fig. \ref{fig:arch}, the overall architecture of our proposed ITTR is consisted of three parts: 1) the stem with three convolution layers for overlapping patch embedding with down-sampling operations; 2) the body of ITTR consisted of 9 hybrid perception blocks stacked in series; 3) the decoder of ITTR which is the mirror of the stem. The design motivation is described below.

Firstly, to enlarge the patch size, we have used stacked convolution layers for overlapping patch embedding. Compared with non-overlapping patch embedding, patch size for overlapping patch embedding is increased to $ 13 \times 13 $ pixels instead of $4 \times 4$. Besides, it is more efficient to decouple overlapping patch embedding layer into three convolution layers. Secondly, to ensure the generator capacity and make a fair comparison, we have stacked 9 HPBs in the body of ITTR so that the number of blocks is the same as the generator in our baseline CUT \cite{cut}. Thirdly, the decoder of ITTR is also implemented with three convolution layers. Rather than a HPB, a convolution layer is more efficient to deal with the high-resolution feature map after up-sampling.


The procedure mentioned above with an input image $I \in \mathbb{R}^{H \times W \times 3}$ can be described as below,

\begin{equation}
  X = \text{Stem}(I), \, X \in \mathbb{R}^{H/4 \times W/4 \times C},
  \label{eq:stem_down}
\end{equation}

\begin{equation}
  X = \text{HPB}_i(X), \, i \in \{1,2...9\}, \, X \in \mathbb{R}^{H/4 \times W/4 \times C},
  \label{eq:body}
\end{equation}

\begin{equation}
  Y = \text{Decoder}(X), \, Y \in \mathbb{R}^{H \times W \times 3}.
  \label{eq:i_out}
\end{equation}

\subsection{Hybrid Perception Block}



Hybrid perception block (HPB) is proposed to capture contextual information from both local perception and global perception.
To achieve this, HPB makes spatial token mixing between adjacent or distant tokens with two parallel branches as shown in Fig. \ref{fig:arch}.
The local branch adopts a convolution layer for its local efficiency.
More precisely, we have used a depth-wise convolution layer for further complexity reduction.
Besides, the global branch adopts self-attention for long-range dependency.
Because the attention map within self-attention establishes relationships between each token pair.

Moreover, channel-wise fusion is expected. An MLP layer is first employed for fusion between positional corresponding tokens extracted by the two branches, which aims to fuse contextual information extracted from different perceptions. Then, a convolutional FFN is adopted for more detailed fusion with channel expansion and reduction. Besides, a depth-wise convolution layer is also used for spatial token mixing after channel expansion. In addition, instance normalization \cite{in,in2} and GELU \cite{gelu} activation are used to stabilize distribution and accelerate convergence.

\begin{figure}[t]
  \centering
  \includegraphics[width=\linewidth]{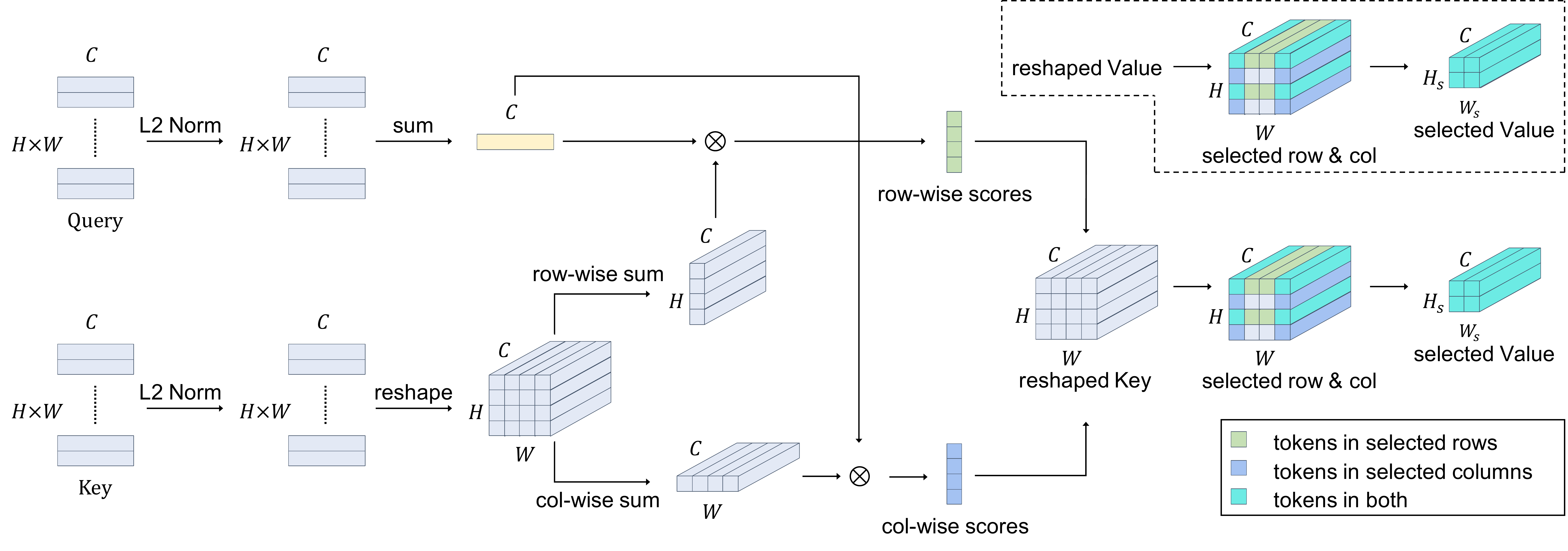}

  \caption{
    Schematic diagram of token pre-pruning mechanism in DPSA as described in Section \ref{sec:dpsa}.
    Tokens in Key are grouped by rows and columns to compute contribution score. 
    Tokens in rows and columns with less contribution are pruned from Key and Value.
    }

  \label{fig:dpsa}
\end{figure}

\subsection{Dual Pruned Self-Attention}
\label{sec:dpsa}
To simplify the description, we take single-head DPSA to explain its mechanism in this subsection.
The forward propagation process of DPSA shown in Fig. \ref{fig:dpsa} can be decomposed into token contribution measurement and token pruning. 


In DPSA, contribution of token $k_j $ in $ K \in \mathbb{R}^{N \times C}$ is defined as $\sum_{i=1}^N{a_{ij}}$.
According to the definition, the computation of the attention map should precede the computation of token contribution.
However, this is unacceptable because the motivation to design DPSA is to pre-prune Key and Value before the computation of the attention map.
Instead, we calculate the contribution of tokens grouped by rows or columns. 
Therefore, the cost of the contribution measurement can be sharply reduced thanks to the distributive property of vector inner product.
Formula of the contribution measurement is displayed bellow, $q_{i}$ and $k_{j}$ are tokens in Query ($Q \in \mathbb{R}^{N \times C}$) and reshaped Key ($K^{'} \in \mathbb{R}^{H \times W \times C} $):
\begin{equation}
     \text{Score}_{r} = \sum\nolimits_{i=0}^{N}\sum\nolimits_{j=0}^{W} q_{i}k_{rj}^{T}  = (\sum\nolimits_{i=0}^{N}q_{i}) (\sum\nolimits_{j=0}^{W} k_{rj})^{T} , \, r \in \{1...H\},
    \label{eq:score_r_1}
\end{equation}
\begin{equation}
    \text{Score}_{c} = \sum\nolimits_{i=0}^{N}\sum\nolimits_{j=0}^{H} q_{i}k_{jc}^{T} =  (\sum\nolimits_{i=0}^{N}q_{i})(\sum\nolimits_{j=0}^{H} k_{jc})^{T} ,\, c \in \{1...W\}.
    \label{eq:score_c_1}
\end{equation}

Notably, the contribution of grouped tokens can be calculated in this way only if token-wise L2 normalization is adopted for Query and Key. 
Since the norm of a token vector in $Q$ or $K$ is normalized to 1, element values in the attention map are restricted to the range $(-1,1)$.
By which, the negative impact of peaked token vectors can be eliminated before Softmax activation.

The computation of contribution score enables token pruning for rows and columns.
We rank among rows and columns by referring to their contribution scores $\text{Score}_r \in \mathbb{R}^{H}$ and $\text{Score}_c \in \mathbb{R}^{W}$ .
Then, indexes of rows or columns with higher contribution scores are selected.
Only tokens in the selected rows and columns are remained while others are pruned.
In addition, the number of selected rows or columns $N_{s}$ is a hyper-parameter, which has been set to the square root of $H$ in our experiments. 
These procedures can be described as following,
\begin{equation}
    \begin{aligned}
            & \text{Index}_{r} = \text{ArgMaxScore}(\text{Score}_{r})[:N_{s}], \\
            & \text{Index}_{c} = \text{ArgMaxScore}(\text{Score}_{c})[:N_{s}],
       \end{aligned}
    \label{eq:select}
\end{equation}

\begin{equation}
    K_s = K^{'}[\text{Index}_{r},\,\text{Index}_{c}], \quad
    V_s = V^{'}[\text{Index}_{r}, \, \text{Index}_{c}].
    \label{eq:selectkv}
\end{equation}



Here, ArgMaxScore is to rank indexes of rows or columns by referring to their contribution scores. Operation $[:N_{s}]$ is to select indexes with top $N_{s}$ ranking.
Then we use these indexes to select tokens in $K^{'}$ and $V^{'} \in \mathbb{R}^{H \times W \times C}$. 
After pruning, selected tokens are reshaped from $K_{s} \in \mathbb{R}^{N_{s} \times N_{s} \times C}$ and $V_{s} \in \mathbb{R}^{N_{s} \times N_{s} \times C}$ to $K_{s} \in \mathbb{R}^{N_{s}^2  \times C}$ and $V_{s} \in \mathbb{R}^{N_{s}^2  \times C}$. 
The subsequent calculation process is similar to the original self-attention. As a by-product, the temperature factor $1 / \sqrt{D_{h}}$ is no longer needed because the pruning of tokens has indirectly element values in rows of $A_\text{cos}$. 
To this end, the DPSA is computed as following,


\begin{equation}
  \text{DPSA}(X) = \text{concat}_{i=1}^{N_{h}}[\text{SparseAttention}_i(X_{i})]W,
  \label{eq:dpsa}
\end{equation}

\begin{equation}
  \text{SparseAttention}(X) = [\text{Softmax}(QK_s^{T})]V_s.
  \label{eq:sparseatt}
\end{equation}

In a single head, the computational complexity of $QK_s^{T}$ and $A_\text{cos}V_{s}$ are reduced to $\mathcal{O}(N N_{s}^{2} C)$. Memory space complexity of DPSA is reduced to $\mathcal{O}(N N_{s}^2)$. Here, $N$ is equal to $H \times W$. The computational cost of contribution score is relatively negligible that can be ignored. Since $N_{s}$ has been set to $\sqrt{H}$ in practice, the overall computational complexity and memory space complexity for a single head of DPSA are reduced to $\mathcal{O}(N H C)$ and $\mathcal{O}(N H)$.

\subsection{Objectives}
To make comparison with the state-of-art methods, we can choose CUT \cite{cut} and LSeSim \cite{lsesim} as the baselines and replace the generator architecture with ITTR. In particular, since CUT is employed in most of our experiments, we present the objectives of CUT as following,

\begin{equation}
    \mathcal{L}_\text{G} = \mathbb{E}_{x\sim{X}}[(1-D(G(x)))^{2}].
  \label{eq:gan_loss}
\end{equation}

\begin{equation}
     \ell(v,v^{+},v^{-})=-\text{log}\left[ \frac{\text{exp}(v \cdot v^{+} / \tau)}
     {\text{exp}(v \cdot v^{+} / \tau)+ \sum\nolimits_{n=1}^{N} \text{exp}(v \cdot v_{n}^{-} / \tau) }\right ],
  \label{eq:contrastive_function}
\end{equation}

\begin{equation}
      \mathcal{L}_\text{PatchNCE}(G,H,X) = \mathbb{E}_{x\sim{X}}\sum_{l=1}^{L}\sum_{s=1}^{S_{l}}\ell(\hat{z}_{l}^{s},z_{l}^{s},z_{l}^{S/s}),
  \label{eq:patchnce}
\end{equation}

\begin{equation}
    \begin{aligned}
        \mathcal{L} = \mathcal{L}_\text{G}(G,D,X,Y) 
        &+\lambda_{X}\mathcal{L}_\text{PatchNCE}(G,H,X)\\ 
        &+\lambda_{Y}\mathcal{L}_\text{PatchNCE}(G,H,Y),
    \end{aligned}
  \label{eq:cut_loss}
\end{equation}

\begin{equation}
  \mathcal{L}_\text{D} = \mathbb{E}_{y\sim{Y}}[(1-D(y))^{2}]+\mathbb{E}_{x\sim{X}}[D(G(x))^{2}].
  \label{eq:dis_loss}
\end{equation}

Here, $X$ and $Y$ are real images from the source domain and the target domain. $G$ and $D$ are generator and discriminator. 
$H_l$ is a two-layer MLP network corresponding to chosen layer $\ell$ in the generator.
$s \in S$ indicates position. $z_{l}^{s}$ and $z_{l}^{S/s}$ are produced by $H_{l}(G_{l}(X))$ but in different positions. $\hat{z}_{l}^{s}$ is produced by $H_{l}(G_{l}(G(X)))$ and has the same position with $z_{l}^{s}$. $\lambda_{X}$ and $\lambda_{Y}$ are hyper-parameters, both of them have been set to 1 in experiments.



\begin{figure}[!t]
  \centering
  \includegraphics[width=\linewidth]{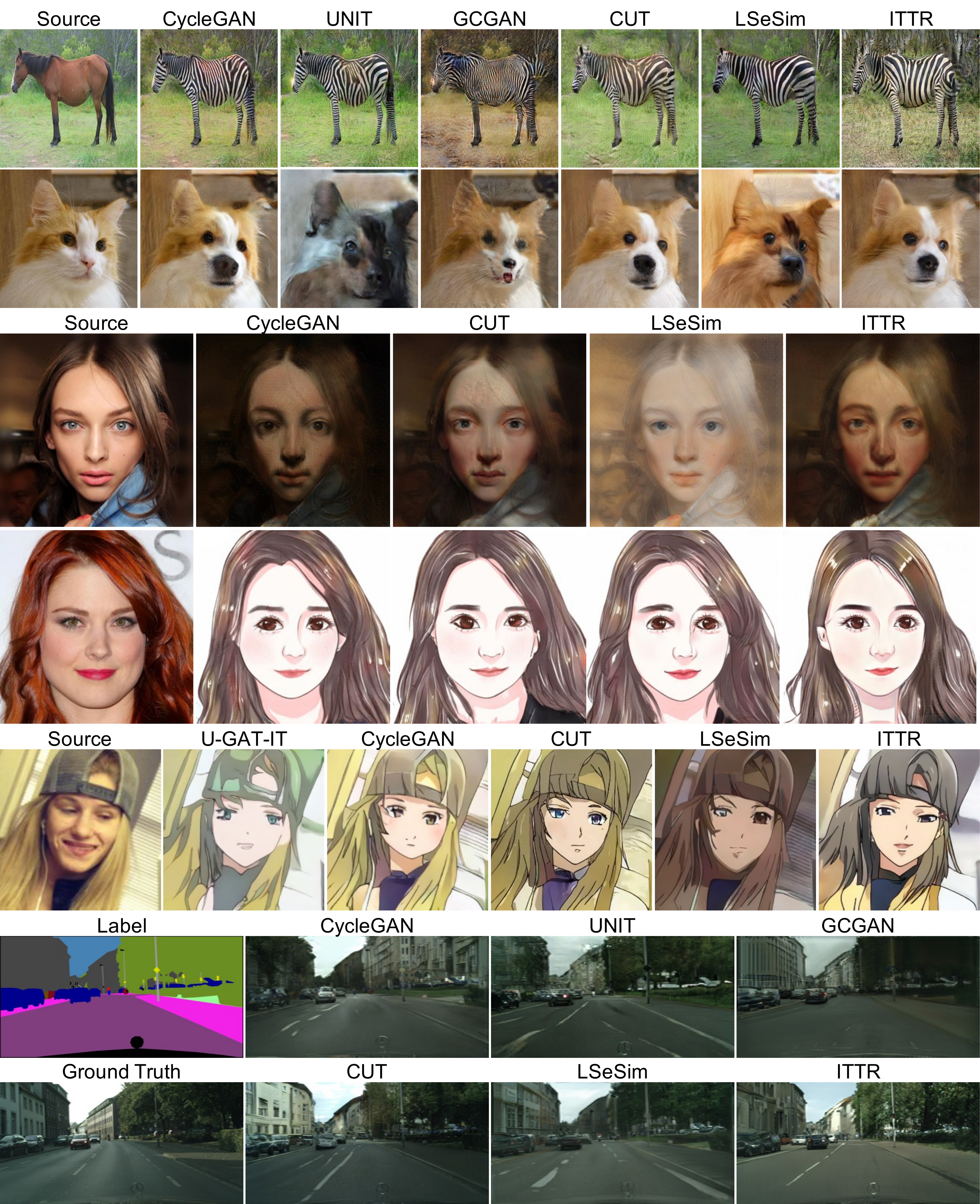}

  \caption{
     Qualitative comparison of results produced by different methods.
  }
  \label{fig:sotas}
\end{figure}
\begin{table}[t]
    \scriptsize
  \centering
   \caption{
    Comparison between ITTR and state-of-arts on three datasets. Evaluation metric with sign $\downarrow$ indicates that lower is better, while $\uparrow$ indicates higher is better. 
    The first three groups on top are published results from CUT \cite{cut} and LSeSim \cite{lsesim}. 
    Results of method name marked by $^{*}$ are our reproductions using publicly-available source code. 
    The bottom two lines are ITTR trained by strategies from CUT and LSeSim. 
  }
      \begin{tabular}{l:cccc:c:c}
        \toprule
        \multirow{2}{*}[-0.5ex]{\textbf{Method}}  & \multicolumn{4}{c}{\textbf{Cityscapes}} & \textbf{Cat $\rightarrow$ Dog} & \textbf{H $\rightarrow$ Z} \\ 
        
        \cmidrule(lr){2-5}  
        \cmidrule(lr){6-6}  
        \cmidrule(lr){7-7}  
        
        &mAP$\uparrow$ &pixAcc$\uparrow$ &clsAcc$\uparrow$ & FID$\downarrow$ & FID$\downarrow$ & FID$\downarrow$ \\
        
        \midrule                       
        CycleGAN \cite{cyclegan}        &20.4 & 57.2 &25.4  & 76.3 & 85.9      & 77.2  \\
        UNIT \cite{unit}                &16.9 & 58.4 &22.5 & 91.4 &104.4      & 98.0 \\ 
        DRIT++ \cite{drit++}            &17.0 & 60.3 &22.2 & 96.2 &123.4      & 88.5 \\ 
        
        \hdashline 
        Distance \cite{distancegan}     &8.4 & 47.2 &12.6 &75.9 &155.3      & 67.2  \\  
        GcGAN \cite{gcgan}              &21.2 & 65.5 &26.6 &57.4 & 96.6      & 86.7 \\
        
        \hdashline  
        CUT \cite{cut}                  &24.7  & 68.8 &30.7 & 56.4 & 76.2      & 45.5 \\
        LSeSim \cite{lsesim}            &$-$ &73.2  &$-$ &49.7 &$-$   & 38.0 \\
        
        \midrule 
        I2ISA$^{*}$ \cite{i2i_sa}   &20.0 &69.5  &25.8 &151.8 &103.6 &248.2 \\

        \hdashline                       
        CycleGAN$^{*}$ \cite{cyclegan}      &22.1 &67.8  &28.7 &65.1 &87.6 &77.2 \\ 
        CUT$^{*}$ \cite{cut}      &29.0 &83.7  &35.8 &47.8 &74.4 &36.4 \\ 
        LSeSim$^{*}$ \cite{lsesim}      &28.9 &75.7  &37.4 &55.9 &72.8 &38.9 \\ 
        
        \hdashline 
        ITTR (CUT)                   &\textbf{32.5} &\textbf{86.0}  &\textbf{39.7} &\textbf{45.1}  &\textbf{68.6}
                        &\textbf{33.6} \\ 
        ITTR (LSeSim)                   &28.9 &78.8 &36.7  &53.3  &68.7  &36.5 \\ 

        \bottomrule
    \end{tabular}
 
  \label{tab:sotas}
\end{table}

\section{Experiment}
\subsection{Experiment Setup}
\subsubsection{Datasets.} 
To demonstrate the effectiveness of ITTR, six benchmark datasets have been selected for training and testing. 
Horse $\rightarrow$ Zebra has 1067 horse images and 1334 zebra images for training, 120 horse images and 140 zebra images for testing. 
Cityscapes has 2975 and 500 street scene images and semantic segmentation labels for training and testing respectively. 
Similar with semantic image synthesis \cite{pix2pixhd,spade}, we translate segmentation labels to scene images.
Cat $\rightarrow$ Dog is a subset of the AFHQ dataset. 5153 cat images and 4739 dog images are used for training. 500 images for each are used for testing. 
Selfie $\rightarrow$ Anime \cite{ugatit} has 3400 selfies and anime images for training. 100 images for each are used for testing. 
Face $\rightarrow$ Metface \cite{metface} has 1500 real face images selected from CelebA \cite{stargan2} and 1336 face images extracted from art. 300 images for each are used for testing. 
Having been sorted by file names, the first 750 males images and 750 females images in CelebA are selected for training and the following 150 males images and 150 females images are selected for testing.
Female $\rightarrow$ Cartoon \cite{cartoon} has 1500 female images from CelebA \cite{stargan2} and 194 cartoon images for training. 300 female images and 10 cartoon images for testing. These female images are also selected after sorting by file names.

\subsubsection{Implementation Details.} 
To make fair comparisons, we adopt the same training frame as our baseline CUT \cite{cut}. PatchGAN \cite{cyclegan} is adopted as our discriminator. Input image resolution is fixed to $256 \times 256$ by resizing the original images.  The learning rate is set to 2e-4 in the first 200 epochs, then linearly reduced to zero in the next 200 epochs. The whole training process lasts 400 epochs in total. Adam \cite{adam} optimizer is employed to update the weights of the network. For each training, only one GPU is used and batch size is set to 1. 
All the hyper-parameters set for training and loss functions are maintained the same as our baseline \cite{cut}. The whole framework is implemented on Pytorch. Experiments are performed on NVIDIA GeForce RTX 2080Ti.

\subsubsection{Evaluation Metrics.} 
We use FID (Fréchet Inception Distance) \cite{pytorchfid,fid} to measure similarity between generated fake images and real images in target domain. 
\begin{equation}
    \text{FID}(Y,\hat{Y})=||\mu_{Y}-\mu_{\hat{Y}}||_{2}^{2} + 
    \text{Tr}({\rm \Sigma}_{Y}+{\rm \Sigma}_{\hat{Y}}-2({\rm \Sigma}_{Y}{\rm \Sigma}_{\hat{Y}})^{\frac{1}{2}}).
  \label{eq:fid}
\end{equation}
Here, $Y$ and $\hat{Y}$ represent the real images and generated images in the target domain. $\mu$ and $\Sigma$ are the mean and covariance of the image set. Moreover, following \cite{lsesim}, we have also used pre-trained DRN \cite{drn} to calculate DRN-Score \cite{pix2pix} for Cityscapes. 
DRN-Score between segmentation result with generated image and ground-truth is calculated. To make a fair comparison, this DRN is trained in the same setting as our baseline. We train DRN using the default recommended setting by publicly-available open-source code \cite{drn}, except the input image resolution is fixed to $256\times256$, which is consistent with our model output.

\begin{table}[t]
\scriptsize
  \centering
    \caption{
      Quantitative experimental results on three face stylization datasets.
  }
      \begin{tabular}{l:c:c:c}
        \toprule
        \multirow{2}{*}[-0.5ex]{\textbf{Method}}  
        & \textbf{Selfie $\rightarrow$ Anime} 
        & \textbf{Face $\rightarrow$ Metface} 
        & \textbf{Female $\rightarrow$ Cartoon}  \\
        
        \cmidrule(lr){2-2}  
        \cmidrule(lr){3-3}  
        \cmidrule(lr){4-4} 
        & FID$\downarrow$ & FID$\downarrow$ & FID$\downarrow$ \\
        
        \midrule
        U-GAT-IT \cite{ugatit}            &85.4  &$-$      &$-$   \\
        CycleGAN \cite{cyclegan}        &84.5   &115.6    &96.2   \\
        CUT \cite{cut}.                 &75.6  &100.9     &93.4   \\
        LSeSim \cite{lsesim}.           &84.5   &112.6    &93.4   \\
        
        \hdashline
        ITTR (CUT).                     &\textbf{73.4}   &\textbf{93.7}  &\textbf{91.6} \\ 

        \bottomrule
    \end{tabular}

  \label{tab:fancys}
\end{table}

\subsection{Comparison with the State-of-the-art Methods}
Image translation results produced by ITTR are displayed in Fig. \ref{fig:demos}. 
Qualitative comparison on results of ITTR and other methods is shown in Fig. \ref{fig:sotas}. 
It is obvious that our method has achieved comparable or better performance visually.
Quantitative evaluation metrics are listed in Table \ref{tab:sotas} and Table \ref{tab:fancys}. 
We regard FID as the reference of the style discrepancies between translation results and real images in the target domain, while DRN-Score is more related to the preservation of content details in the source image.
ITTR has achieved better performance on both metrics over six datasets, which indicates that our method outperforms others with a better ability to translate an image into the target domain while maintaining content details.

Besides, statistics of generator MACs and parameters for each method are shown in Table \ref{tab:comlexity}. Although ITTR is not the most lightweight one, it is still efficient enough to achieve better performance with less complexity than the most competitive methods CUT \cite{cut} and LSeSim \cite{lsesim}.

\begin{table}[t]
    \scriptsize
  \centering
    \caption{
  Statistics of generator MACs and parameters for each method.
  }
      \begin{tabular}{l:c:c}
        \toprule
        \textbf{Method} & \textbf{MACs (G)} & \textbf{Params (M)} \\
    
        \midrule                  
        I2ISA \cite{i2i_sa}     &85.7 &15.1 \\  
        
        \hdashline
        CycleGAN \cite{cyclegan}        &56.8 &11.4 \\
        UNIT \cite{unit}                &71.1 & 11.1\\ 
        DRIT++ \cite{drit++}            &119.4 &8.6\\ 
        
        \hdashline 
        Distance \cite{distancegan}     &8.3 &5.5 \\  
        GcGAN \cite{gcgan}              &42.3 &7.8\\
        
        \hdashline  
        CUT \cite{cut}                  &64.1 &11.4\\
        LSeSim \cite{lsesim}            &64.1 &11.4\\

        \hdashline 
        ITTR (CUT)          &45.8 &8.5\\ 
        \bottomrule
    \end{tabular}
  \label{tab:comlexity}
\end{table}



\begin{table}[t]
\scriptsize
  \centering
    \caption{ 
    Quantitative experimental results for ablation study.
  }
      \begin{tabular}{l:l:cccc:c}
        \toprule
        \multirow{8}{*}[-0.5ex]{} &
        \multirow{2}{*}[-0.5ex]{\textbf{Configuration}} & \multicolumn{4}{c}{\textbf{Cityscapes}} & \textbf{H $\rightarrow$ Z}  \\ 
        
        \cmidrule(lr){3-6}  \cmidrule(lr){7-7}  
        & &mAP$\uparrow$ &pixAcc$\uparrow$ &clsAcc$\uparrow$ & FID$\downarrow$  & FID$\downarrow$ \\

        \midrule
        
        A & w/o Local Perception                              &30.5 &84.2 &37.3 &45.7  &40.0 \\
        B & w/o DPSA                                    &29.6 &83.7 &36.9 &48.6  &35.6 \\
        \hdashline
        C & DPSA $\rightarrow$ Spatial Reduction Att   &29.7 &84.3 &36.5 &45.8  &36.0 \\
        D & DPSA $\rightarrow$ Global Sparse Att       &32.1 &85.7 &39.2 &46.4  &37.7 \\
        \hdashline
        E & w/o L2 Norm      &2.1 &20.4 &5.8 &388.1  &416.5 \\ 
        \hdashline 
        F & Ours &\textbf{32.5} &\textbf{86.0}  &\textbf{39.7} &\textbf{45.1} &\textbf{33.6}\\

        \bottomrule
    \end{tabular}
  \label{tab:ablation}
\end{table}

\begin{figure}[b]
  \centering
  \includegraphics[width=\linewidth]{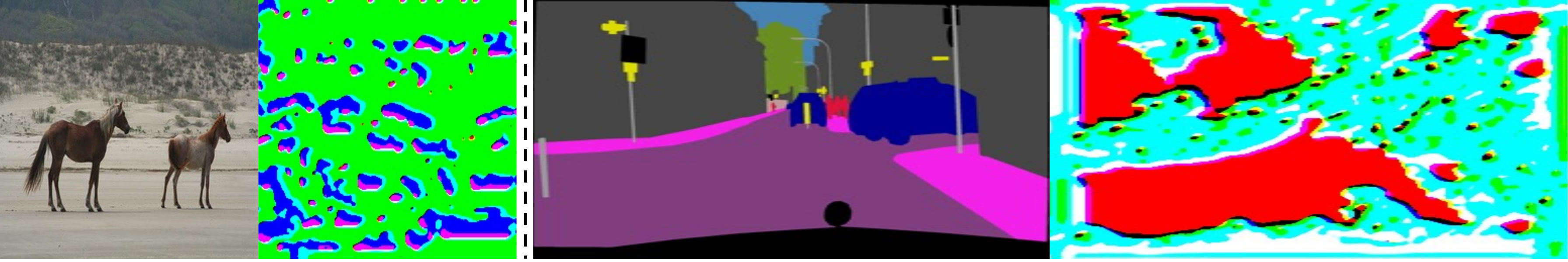}

  \caption{Translation results produced by a collapsed generator without token-wise L2 Normalization for $Q$ and $K$.}
  \label{fig:collapsed}
\end{figure}

\subsection{Ablation Study}

Experimental results for ablation study are listed in Table \ref{tab:ablation}. 
Firstly, we has alternatively removed one of the branches in HPB. 
Comparison between config A and B has shown that FID is worse without a local perception branch, while DRN-Score is worse without DPSA. 
This phenomenon indicates that DPSA for long-range contextual information is more concentrated on content details and the depth-wise convolution for local perception is beneficial to eliminate gaps between domains.
Secondly, to examine the necessity of using DPSA, config C has replaced DPSA by spatial reduction attention (SRA) \cite{pvtv1}, which is one of the most efficient designs for ViTs. 
Thirdly, config D has computed the contribution score for each token instead of row-wise and column-wise token groups.
Experimental results of Config C and D have verified the superiority and effectiveness of DPSA. 
Finally, in config E, we have removed the token-wise L2 Normalization for $Q$ and $K$, and the model has collapsed as shown in Fig. \ref{fig:collapsed}, which has also verified the argumentation in Section \ref{sec:dpsa}.

\begin{table}[t]
\scriptsize
  \centering
    \caption{
  Comparison of DPSA with different numbers of sparse tokens.
  Experiments are made with number of sparse tokens from $1 \times 1$ to $ 16 \times 16 $.
  }
      \begin{tabular}{c:cccc:c}
        \toprule
        \multirow{2}{*}[-0.5ex]{\textbf{Tokens}} & \multicolumn{4}{c}{\textbf{Cityscapes}} & \textbf{H $\rightarrow$ Z}  \\ 
        
        \cmidrule(lr){2-5}  \cmidrule(lr){6-6}  
        &mAP$\uparrow$ &pixAcc$\uparrow$ &clsAcc$\uparrow$ & FID$\downarrow$  & FID$\downarrow$ \\

        \midrule
        $1 \times 1$                            &31.3 &85.5 &37.9 &45.4  &35.1 \\
        $2 \times 2$                             &31.9 &85.6 &39.1 &45.8  &35.0 \\
        $4 \times 4$                              &30.9 &85.6 &37.6 &45.5  &36.1 \\
        $8 \times 8$   &\textbf{32.5} &\textbf{86.0} &\textbf{39.7} &\textbf{45.1}  &\textbf{33.6} \\
        $16 \times 16$                      &31.5 &85.2 &38.7 &46.8  &36.2 \\
        \bottomrule
    \end{tabular}
  \label{tab:ablation2}
\end{table}

\subsection{Number of Sparse Tokens }
We have evaluated the influence caused by the number of selected tokens in DPSA as shown in Table \ref{tab:ablation2}. 
It is reasonable that DPSAs with the number of sparse tokens less than $8 \times 8$ have made worse performance because of the reduction in parameters. 
However, the experiment with $16 \times 16$ sparse tokens has made the worst translation performance. 
After analysis, we have found this is related to the Softmax activation. 
In DPSA, we have removed the scale factor because the Softmax can be smoothed by token-wise L2 Normalization and token pruning.
But the increment of selected tokens can soften the impact of token pruning, which results in hardness of convergence and finally leads to bad image-to-image translation performance after training.



\section{Conclusions}
In this paper, we aim to propose an effective and efficient architecture generator architecture for unpaired image-to-image translation.
To achieve this, we have proposed ITTR, which has integrated transformer-based architecture to capture long-range dependency.
In order to capture contextual information from different perceptions, we have designed the hybrid perception block (HPB). 
To reduce computational complexity and memory space complexity, we have presented the dual pruned self-attention (DPSA).
Experiments on six datasets have verified the generalization ability of our method. 
Visualization of Grad-CAM has shown that our transformer-based generator can capture more instance-level contextual information than CNN-based generator by long-range dependency capturing.
Qualitative and quantitative comparison have shown that our method achieved better performance to the state-of-the-arts.



\clearpage
%
%
\bibliographystyle{splncs04}
\bibliography{egbib}
\end{document}